\documentclass[conference]{IEEEtran}
\IEEEoverridecommandlockouts
\usepackage{cite}
\usepackage{amsmath,amssymb,amsfonts}
\usepackage{algorithmic}
\usepackage{graphicx}
\usepackage{textcomp}
\usepackage{xcolor}
\usepackage{tabularx}
\usepackage{booktabs}
\usepackage[export]{adjustbox}
\usepackage{tcolorbox}
\usepackage{hyperref}
\usepackage{csquotes} 

\def\BibTeX{{\rm B\kern-.05em{\sc i\kern-.025em b}\kern-.08em
    T\kern-.1667em\lower.7ex\hbox{E}\kern-.125emX}}
 
\begin{document}




\title{[Vision Paper] PRObot: Enhancing Patient-Reported Outcome Measures for Diabetic Retinopathy using Chatbots and Generative AI}

\author{
    \IEEEauthorblockN{Maren Pielka\IEEEauthorrefmark{1}\IEEEauthorrefmark{2}\IEEEauthorrefmark{4},
    Tobias Schneider\IEEEauthorrefmark{1}\IEEEauthorrefmark{2}\IEEEauthorrefmark{4}, 
    Jan Terheyden\IEEEauthorrefmark{3},
    Rafet Sifa\IEEEauthorrefmark{1}\IEEEauthorrefmark{2}
    }
    \IEEEauthorblockA{\IEEEauthorrefmark{1}Fraunhofer IAIS, Sankt Augustin, Germany}
    \IEEEauthorblockA{\IEEEauthorrefmark{2}University of Bonn, Bonn, Germany}
    \IEEEauthorblockA{\IEEEauthorrefmark{3}Universitätsklinikum Bonn, Bonn, Germany}
    \IEEEauthorblockA{\IEEEauthorrefmark{4}Equal contribution}
    \texttt{Maren.Pielka@iais.fraunhofer.de}
}

\maketitle \footnote{This work was accepted into the proceedings of the IEEE Big Data 2024 conference. The copyright lies with IEEE.}

\begin{abstract}
We present an outline of the first large language model (LLM) based chatbot application in the context of patient-reported outcome measures (PROMs) for diabetic retinopathy. By utilizing the capabilities of current LLMs, we enable patients to provide feedback about their quality of life and treatment progress via an interactive application. The proposed framework offers significant advantages over the current approach, which encompasses only qualitative collection of survey data or a static survey with limited answer options. Using the PROBot LLM-PROM application, patients will be asked tailored questions about their individual challenges, and can give more detailed feedback on the progress of their treatment. Based on this input, we will use machine learning to infer conventional PROM scores, which can be used by clinicians to evaluate the treatment status. The goal of the application is to improve adherence to the healthcare system and treatments, and thus ultimately reduce cases of subsequent vision impairment. The approach needs to be further validated using a survey and a clinical study.
\end{abstract}

\begin{IEEEkeywords}
Machine Learning, Large Language Models, Natural Language Understanding, Patient-Reported Outcome Measures, Ophthalmology
\end{IEEEkeywords}

\section{Introduction}
Monitoring chronic eye diseases is increasingly based on data from imaging exams, functional tests and self-reports of patients. While treatment decisions in medicine depend on all three columns, quantitative data are often available only for imaging biomarkers and functional assessments, while symptoms and health behaviors are mostly assessed qualitatively. Structured questionnaires such as  patient-reported outcome measures (PROMs), e.g. the National Eye Institute Visual Function Questionnaire 25 (NEI-VFQ-25, table \ref{tab:nei_vfq_examples}) \cite{10.1001/archopht.116.11.1496, 10.1001/archopht.119.7.1050}, are increasingly used to foster patient-centered care and provide quantitative metrics of patient-reports to clinicians. However, they have been developed in the context of clinical trials and are therefore highly static and not necessarily user-friendly.
In PROM questionnaires, patients are asked questions e.g. about their physical and mental well-being, which are to be answered with a score on a numerical scale.

\begin{table}[h]
    \centering
    \begin{tabularx}{\linewidth}{c} \toprule
        \textbf{Question} \\ \midrule 
        \begin{minipage}[r]{0.95\linewidth} How much of the time do you worry about your eyesight? \end{minipage}  \\ \midrule 
        \begin{minipage}[r]{0.95\linewidth} At the present time, would you say your eyesight using both eyes (with glasses or contact lenses, if you wear them) is excellent, good, fair, poor, or very poor or are you completely blind?
 \end{minipage} \\  \midrule 
        \begin{minipage}[r]{0.95\linewidth} How much difficulty do you have driving in difficult conditions, such
as in bad weather, during rush hour, on the freeway, or in city traffic?
 \end{minipage}  
      \\ \midrule
       \begin{minipage}[r]{0.95\linewidth} How much pain or discomfort have you had in and around your eyes
(for example, burning, itching, or aching)?
\end{minipage}
        \\ \midrule
        \begin{minipage}[r]{0.95\linewidth}How much difficulty do you have reading ordinary print in
newspapers? 
\end{minipage}
        \\ \midrule
        \begin{minipage}[r]{0.95\linewidth}Because of your eyesight, how much difficulty do you have visiting
with people in their homes, at parties, or in restaurants ?
\end{minipage} \\
        \bottomrule \vspace{1mm}
    \end{tabularx}
    \caption{Exemplary questions from the NEI-VFQ-25 questionnaire for eye disease patients.}
    \label{tab:nei_vfq_examples}
\end{table}

Due to the static and repetitive design of PROM questionnaires with little to no individualization, low adherence and loss to follow-up are common themes in real-world applications of PROMs in the context of chronic conditions. This impedes the high degree of patient empowerment that can be reached by implementing PROMs into health services. This issue becomes particularly relevant in the case of diabetic retinopathy, which is a prominent and increasing cause of global blindness. Advanced stages of diabetic retinopathy require invasive treatments such as regular injections into the eye, laser coagulation or surgery, which prevents blindness in the affected patients. However, a considerable proportion of people with diabetic retinopathy skips appointments or cancels the treatment altogether, which suggests that the communication between patients and care providers needs to be improved and patients need to be empowered at an earlier stage. 

Given this status quo, and considering the impressive development of generative artificial intelligence (GenAI), and especially large language models (LLMs) in the last years, we came up with the vision of an interactive chatbot to replace the static PROM-questionnaires. Instead of lacking communication between patients and care providers or a repetitive set of questions, patients can \enquote{speak} to the chatbot in real time and answer questions that are streamlined to their personal situation and background. In order to obtain numeric scores for measuring the progress of the treatment - similar to the conventional PROM framework -, another machine learning (ML) model is being trained to predict those based on the patient's responses. This approach is being validated by comparing the scores predicted by the ML model to those obtained from the original PROM questionnaire. We aim to cooperate with health providers and make our research available to patients via an application. According to the idea of merging the PROM framework with a chatbot interface, we call our approach \enquote{PRObot}.

In the scope of this vision paper, we qualitatively evaluate the LLM approach, using synthetic data and chatbot interactions with GPT-4o \cite{OpenAI23}. The next steps involve setting up a survey, followed by a clinical study to get both simulated and actual patient data in order to train the ML interpreter model and quantitatively evaluate the results.

\section{Related Work}
Traditionally, medical questionnaires such as PROMs are designed according to Likert-type scales \cite{zis-Likert1932A}. This is a standard approach for surveys where the answers are to be given on a discrete scale. In this framework, single questions are referred to as "items", and the answer options usually correspond to the level of agreement or disagreement with a given statement.

There are a number of possible applications for artificial intelligence and machine learning in medicine, as well as ophthalmology specifically. The majority are image-based methods with a focus on detecting and classifying abnormalities \cite{biesner2022, schneider2024the}. Using AI for interaction with patients - specifically PROMs - is a relatively under-developed area. There has been some previous work on using the outcome of PROMs to improve the therapy for certain diseases \cite{lewy2021}. Improving or replacing the PROMs themselves by a generative AI approach has, to our best knowledge, not been attempted yet.

The fast development in the field of transformer-based language models \cite{vaswani_attention_2017}, especially large generative pre-trained transformers  (GPT) \cite{radford2018improving, openai_gpt3, OpenAI23}, has given rise to new possibilities for language modeling and conversational AI. While the most potent models such as GPT-4o are only accessible via commercial APIs, there are also a number of open-source solutions that achieve competitive results, such as Mixtral \cite{jiang2024mixtralexperts}, LLaMa3 \cite{touvron2023llamaopenefficientfoundation} or Phi-3 \cite{abdin2024phi3technicalreporthighly}. We aim to leverage the generative power of those models by deploying them in an interactive patient-chatbot setting, while providing background information about the traditional PROM items, as well as the patient's history in the prompt. In addition, we want to fine-tune smaller, encoder based models such as BERT \cite{bert2018} or XMLRoBERTa \cite{xlmroberta2019} for predicting the PROM scores based on the patient's answers.

\section{Methodology}\label{method}


PROMs are established and well-researched instruments for collecting patient-recorded data.
However, due to their static design, PROMs are limited in the information they can capture, as any information not covered by PROM items cannot be recorded.
Designing and validating a PROM questionnaire is a complex and time-consuming iterative process.
But even the items of an established and well-calibrated PROM can have a varying relevance due to the strong differences in humans.
For example, in the NEI-VFQ-25 questionnaire \cite{10.1001/archopht.116.11.1496, 10.1001/archopht.119.7.1050}, the influence of people's eyesight on mobility in everyday life is primarily measured by questions relating to driving a car.
However, these questions might be irrelevant for many individuals. 

We propose a novel PROM framework that leverages the capabilities of LLMs to perform a dynamic and more personalized survey of patients.
Instead of relying on static questions and predefined answer options, this framework is based on an interactive chat environment in which patients can respond with free text.
Answers based on free text can convey much more information, enabling a more refined and personalized evaluation of responses.
Additionally, this framework can perform personalized and dynamic questioning for individual patients, using general information about the patient, previous interviews, and even the current chat history to improve the relevance of questions. 
A chat environment is also capable of responding to the free text answers written by the patients.
For example, a chatbot could react emphatically and optimistically to negative responses.
If a patient reports improvements, a chatbot could react positively to help the patient value positive developments\cite{VeraSorin2023LargeLM}.

Using the NEI-VFQ-25 PROM as an example, our goal is to test out a large variety of machine learning models and approaches to derive an optimal framework for building LLM-based PROMS. 
We will report on the performance and benefits of certain approaches to derive recommendations for LLM-based PROMS and determine an optimal configuration of the PROBot framework.

\subsection{Proposed Framework}

\begin{figure}
    \centering
    \includegraphics[width=0.95\linewidth]{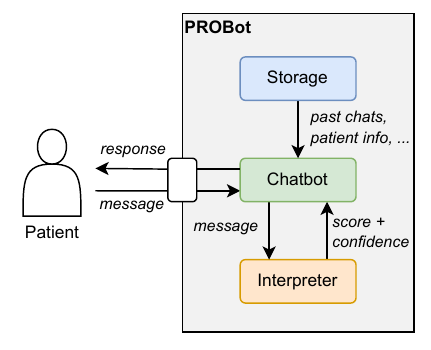}
    \caption{Illustration of the PRObot framework}
    \label{fig:framework}
\end{figure}

The proposed framework consists of the three main components \textit{Interpreter}, \textit{Chatbot} and \textit{Storage}, as visualized in Figure \ref{fig:framework}.
The Interpreter is responsible for understanding and analyzing patient responses.
Based on the results of the Interpreter and data from the Storage component, the Chatbot generates appropriate messages for the dialog with patients.
The \textit{Storage} component provides the available patient data such as past conversations or various patient information.
The Interpreter uses encoder LLMs to extract meaningful vector embeddings of patient responses and the Chatbot uses LLM Prompts and Prompt Engineering to generate high-qualitative and creative text messages.
Both components rely on the strong capabilities of modern LLMs.

The Interpreter consists of an \textit{Encoder} LLM that is used to extract meaningful embeddings for free-text formulations, followed by a machine-learning based \textit{Rater} which derives output scores based on these embeddings.
State-of-the-art Encoder LLMs, such as leading models for the MTEB Benchmark\cite{cao2024recentadvancestextembedding}, will be fine-tuned using task-specific reference text data, which is collected in a survey.
The Encoder LLMs can either be fine-tuned directly on the text data in a self- or unsupervised fashion. Or the entire Interpreter model can be trained end-to-end, which means training the Encoder LLM and Rater together, using the error of output scores as a training loss for the entire model.

Various ML models will be tested to derive output scores, using the NEI-VFQ-25 PROM as a reference.
Different ML architectures such as MLPs or XGBoost will be compared. 
There are different reasonable model outputs, the most widely used being regression or classification: 
While regression considers the order of output classes, the output is continuous and typically there is no measure of confidence.
Classification on the other hand produces discrete outputs. 
Using probabilities as output values can be interpreted as the model's confidence. 
But the ordering of classes is not considered in the training, which can harm the model's performance.
But also more sophisticated approaches, such as Ordinal Classification\cite{Frank2001ASA} which models the order of classes, or Bayesian Regression\cite{Goldstein1976BayesianAO} which models uncertainty for Regression, exist.

Another important design decision is the composition and sharing of components inside the framework architecture. 
The outputs of the model could either be a score for each PROM item, for each PROM subscale or the model could even directly calculate the final aggregated score.
Additionally, the Embedder or Rater can be re-used for different PROM items/subscales or unique components can be trained for each one.
Methods that map a reduced number of free text responses, such as one response per PROM subscale, directly to the subscale scores or the final PROM score have the potential to greatly reduce the required number of patient responses through the strong capabilities of LLMs.
However, this also makes the model less transparent.
Therefore, in addition to performance, various configurations will also be evaluated with regard to explainability, as it is highly relevant for medical applications.

For the Chatbot component of the PROBot framework, powerful capabilities of modern LLMs are used to generate chat messages.  
An important decision when developing the Chatbot component is whether to implement it by training a task-specific LLM or by leveraging LLM Prompting with pre-trained models such as ChatGPT\cite{OpenAI23} or Llama-3\cite{touvron2023llamaopenefficientfoundation}.
Training a new task-specific LLM would require a lot of clean data demonstrating the intended chatbot behaviour.
This kind of data is difficult to obtain and the resulting model would have limited customizability.
However, many pre-trained LLMs such as ChatGPT are intended for chatbot usage and therefore already fulfill many relevant requirements for our framework as LLM alignment is a major priority in the development of such models. 
Relevant aspects such as reliability, safety, fairness, resistance to misuse, explainability \& reasoning, social norms and robustness are considered in LLM alignment\cite{liu2024trustworthyllmssurveyguideline}.
By defining user prompts, LLMs become highly customizable as they allow for a precise description of the chatbot behaviour. 
Therefore, the chatbot's behaviour can also be easily optimized in the future.
A prompt-based approach enables cost-effective reuse of the framework for other use cases as well.
Therefore, we propose to use LLM Prompting for the Chatbot component.

\subsection{Data Collection and Evaluation}

To ensure high quality for the chatbot text generation and interpretation of patient responses, large amounts of clean data need to be collected to fine-tune and evaluate task-specific large language models.
A large corpus of patient surveys will be acquired, based on the NEI-VFQ-25 PROM and free text formulations.
This data allows the training and comparison of many kinds of machine learning models with respect to the broadly accepted NEI-VFQ-25 PROM.
Each participant will fill out the NEI-VFQ-25 PROM, which can then be used for target scores when training and evaluating ML models.

The participants are asked to write free text formulations in different formats.
To ensure that the time required to complete the survey is reasonable, each participant will only complete one of the formats.
Participants may be asked to respond to each individual PROM item with a free text, or to respond to a question for each PROM subscale with a free text.
These formats are useful as they allow a direct mapping from the free text formulations to the target scores of the NEI-VFQ-25 PROM.
Other formats such as writing a single large free-text formulation or using other scales such as 10 quality-of-life domains are also possible.
However, these may be more difficult to map to the target scores of the NEI-VFQ-25 PROM.
In this case, the participants will provide a score for each written free text, which can also be used as a target for the training and evaluation of models.

The collected data will be used to train ML models to learn the mapping from free-text responses to scores such as the individual PROM items or the PROM subscales.
By collecting free-text responses for individual PROM items as well as for PROM subscales, the ability of LLMs to derive meaningful scores for entire subscales with just one text can be analyzed.

To test the chatbot's text generation based on LLM prompting, interactive experiments will be conducted with individuals.
By utilizing evaluation methods such as A/B testing, the performance of different LLM configurations can be compared.
The iterative execution of this approach with an increasing number of participants allows to gradually evaluate and improve the chatbot's text generation capabilities.
Conducting experiments with individuals is an expensive and time-consuming process.
But the findings and data are very important for optimizing the chatbot for interaction with real people.
While it is not reasonable to omit experiments with humans, it has been shown that LLMs can be utilized as strong user simulators \cite{ALGHERAIRY2025101697}.
Therefore, LLM-based user simulators could be used in earlier stages to optimize an initial LLM prompt for the Chatbot, which can be used as a baseline for the first experiments with real users.

The performance of the Interpreter will mainly be evaluated quantitatively by analyzing the error of predicted scores compared to their NEI-VFQ-25 target, using various regression or classification metrics, such as mean squared error, accuracy or F1-Score.
Beside raw model performance, possible biases of the Interpreter such as age- or gender-biases should also be analyzed. 
On the other hand, evaluating the performance of the chatbot is not that straightforward. 
While quantitative evaluation, using explicit metrics such as user feedback or implicit metrics such as conversation length or abortion rate, can be used to gain insights about the quality of the text generation, qualitative analysis still plays an important role when evaluating the outputs for text generation tasks.

\section{Qualitative Results}

\begin{table}[h!]
\begin{tabularx}{\columnwidth}{X}
\toprule
\begin{minipage}{.97\columnwidth} \textbf{PROTOBot}: Hello, I’m PROTOBot, here to support you with your eye condition. To better understand how your vision affects you, could you please tell me a bit about your profession, hobbies, and other activities that might be influenced by your eyesight? \end{minipage} \\ \midrule
\begin{minipage}{.97\columnwidth} \textbf{User}: I’m Michaela Schmidt. As a graphic designer, my blurry vision makes it really hard to work on detailed projects and color selections. I love photography and hiking, but focusing my camera and navigating rough trails have become increasingly difficult. \end{minipage} \\ \midrule
\begin{minipage}{.97\columnwidth} \textbf{PROTOBot}: Thank you, Michaela, for sharing that with me. It sounds like your profession and hobbies rely heavily on good vision. Let's delve a bit deeper to understand how your eye condition impacts different aspects of your life.\\
How often do your vision problems interfere with your ability to work on detailed projects and make accurate color selections? \end{minipage} \\ \midrule
\begin{minipage}{.97\columnwidth} \textbf{User}: My vision problems interfere with my work on detailed projects and color selection almost daily. It has become really hard to distinguish between subtle color differences and focus on fine details. \end{minipage} \\ \midrule
\begin{minipage}{.97\columnwidth} \textbf{PROTOBot}: That must be incredibly challenging, especially in a field where precision is so important. I'm truly sorry you're experiencing this difficulty.\\
How has your enjoyment of photography been affected by your blurry vision, particularly when trying to focus your camera or capture detailed shots? \end{minipage} \\ \midrule
\begin{minipage}{.97\columnwidth} \textbf{User}: It's been quite frustrating. I often struggle with focusing my camera, and capturing detailed shots has become very challenging.  \end{minipage} \\ \midrule
\begin{minipage}{.97\columnwidth} \textbf{PROTOBot}: I can understand how disheartening that must be, given your passion for photography. It’s tough when a loved hobby becomes a source of frustration. \end{minipage} \\ \bottomrule

\vspace{1mm}
\end{tabularx} 
\caption{Excerpt from exemplary dialogue of PROTOBot \\and a simulated user using GPT-4o}
\label{tab:dialog1}
\end{table}

\begin{table}[h!]
\begin{tabularx}{\columnwidth}{X}
\toprule
\begin{minipage}{.97\columnwidth}\textbf{PROTOBot}: Hello, I'm PROTOBot, here to support you with any questions you might have related to your eye health.\\
To better assist you, could you please tell me more about yourself? Specifically, your profession, hobbies, and other activities that may be affected by your eyesight. \end{minipage} \\ \midrule
\begin{minipage}{.97\columnwidth}\textbf{User}: Hello, I'm Sandra Becker. I'm a teacher, which means I spend a lot of time grading student work. I love painting and practicing yoga in my free time. My cataracts make my vision cloudy, so I have to focus on larger, simpler motifs in my artwork, and yoga helps me maintain my balance. \end{minipage} \\ \midrule
\begin{minipage}{.97\columnwidth}\textbf{PROTOBot}: Hi Sandra, thank you for sharing more about yourself. It sounds like your cataracts have quite an impact on your daily life and hobbies.\\
Could you describe how your cataracts affect your ability to read and grade student work? \end{minipage} \\ \midrule
\begin{minipage}{.97\columnwidth}\textbf{User}: Reading and grading student work has become much more challenging due to the cloudiness in my vision. I often have to use brighter lighting and magnifying tools to see the text clearly. \end{minipage} \\ \midrule
\begin{minipage}{.97\columnwidth}\textbf{PROTOBot}: I'm sorry to hear that reading and grading student work has become more difficult for you, Sandra. It's admirable how you are adapting to these challenges to continue your work.\\
How does your vision affect your ability to enjoy painting? Are there specific techniques or tools you've had to change or implement? \end{minipage} \\ \midrule
\begin{minipage}{.97\columnwidth}\textbf{User}: The cloudiness from the cataracts makes it hard to see fine details, so I focus on larger, simpler motifs when I paint. I've also started using brighter colors to make the contrasts more noticeable. \end{minipage} \\ \midrule
\begin{minipage}{.97\columnwidth}\textbf{PROTOBot}: That sounds like a positive approach and a creative way to continue enjoying your painting despite the challenges. It's great to hear that you are finding ways to adapt.\\
Has your vision impacted your ability to practice yoga, particularly with maintaining balance or following along with yoga classes? \end{minipage}\\ \midrule
\begin{minipage}{.97\columnwidth}\textbf{User}: Yes, my vision has made it a bit more challenging to maintain my balance during yoga. However, by focusing on my other senses and practicing regularly, I've managed to adapt and still enjoy my sessions. \end{minipage} \\ \midrule
\begin{minipage}{.97\columnwidth}\textbf{PROTOBot}: It's wonderful to hear that you've found ways to adapt and continue with your yoga practice despite the challenges. Maintaining such a positive attitude is truly inspiring. \end{minipage} \\ \bottomrule

\vspace{1mm}
\end{tabularx}
\caption{Excerpt from exemplary dialogue of PROTOBot \\and a simulated user using GPT-4o}
\label{tab:dialog2}
\end{table}

To test the proposed chat generation approach, we have created a prototype for PRObot, which we call \enquote{PROTObot}.
In order to generate exemplary dialogues for an initial qualitative analysis, we simulate synthetic patients using LLMs. 
In turns, responses are generated for PROTObot and the simulated patient using two separate instances of GPT-4o.
The PROTObot instance is assigned with asking the patient quality-of-life questions regarding their eyesight and address the patients responses to earlier questions with empathy.
For this, the NEI-VFQ-25 questionnaire is provided as reference material.
The instance for the simulated patients is tasked with responding to the messages of the PROTObot instance, while following an LLM generated persona.
Table \ref{tab:dialog1} and \ref{tab:dialog2} show excerpts from generated dialogues with the synthetic patients Michaela Schmidt and Sandra Becker.
The examples show that the simulation of patients results in the generation of exciting dialogues, which enables an initial evaluation of the proposed approach.
One can see that PROTObot is able to generate individually suitable questions for the patients and react appropriately to the patient's responses.
The generated questions are not just copied from the NEI-VFQ-25 PROM, but in most cases highly personalized questions that are tailored to each patient's available information.
However, the scope of the questions remains similar to the reference questions.
In the dialogue with Michaela, PROTObot demonstrates a great amount of empathy. Since she mainly reports on frustrating difficulties in her everyday life, this is very appropriate.
PROTObot also shows a lot of empathy and understanding in the dialogue with Sandra.
In the later messages, Sandra reports on techniques for dealing with her problems more positively.
Here, PROTOBot highlights the positive aspects about these techniques, but without disregarding her underlying issues.
Despite the strong empathy that PROTObot demonstrates, the interactions do not come across as overly solicitous, but rather natural.
However, it should be noted that this can be perceived differently depending on the individual and should be evaluated further.


\section{Conclusion and Outlook}
We presented the outline for an AI-driven approach to improve the quality and usefulness of patient-reported outcome measures for diabetic retinopathy. Next steps involve conducting a survey, as well as a clinical study to collect data for training and validation. On this basis, we want to further refine our approach and make it accessible to health care providers and patients. 

Further development includes validating the ideas developed in this vision paper, i.e. training and evaluating ML models for the chatbot setup and the score prediction. If the approach proves successful for diabethic retinopathy, it could easily be extended to other chronic diseases and prevention use cases, such as oncology, nutrition, or smoking dehabituation. We find that LLMs and chatbots offer great opportunities to help patients as well as health care professionals in understanding and improving factors for treatment success.

\newpage
\section*{Disclosures}
From our point of view, the following conflicts of interest with health care providers are possible: Bayer (Funding), Novartis (Funding, Speaker Fees), Roche (Funding), Okko (Consultancy).

\section*{Acknowledgment}
The project was funded by the Federal Ministry of Education and Research (BMBF) under grant no. 01IS22094A WEST-AI. 

This research has been partially funded by the Federal Ministry of
Education and Research of Germany and the state of North-Rhine
Westphalia as part of the Lamarr-Institute for Machine Learning
and Artificial Intelligence.

\newpage
\bibliography{bibliography}
\bibliographystyle{plain}



\end{document}